%% file: main.tex
\documentclass[10pt,twocolumn,letterpaper]{article}

\usepackage[pagenumbers]{cvpr} %

\input{__misc__}

\def\cvprPaperID{1404} %
\def\confName{CVPR}
\def\confYear{2022}

\begin{document}

\title{ReSTR: Convolution-free Referring Image Segmentation Using Transformers}
\author{
Namyup Kim$^1$ \hspace{4mm} Dongwon Kim$^1$ \hspace{4.5mm} Cuiling Lan$^2$ \hspace{4.5mm} Wenjun Zeng$^3$ \hspace{4.5mm} Suha Kwak$^1$ \hspace{3.5mm}\\
$^1$POSTECH \hspace{8mm} $^2$Microsoft Research Asia \hspace{8mm} $^3$EIT Institute for Advanced Study\\
{\tt\small \url{http://cvlab.postech.ac.kr/research/restr/}}
}

\maketitle

\input{_0_abstract}

\input{_1_introduction}

\input{_2_relatedwork}

\input{_3_method}

\input{_4_experiments}

\input{_5_conclusion}

\clearpage
{\small
\bibliographystyle{ieee_fullname}
\bibliography{cvlab_kwak}
}

\clearpage

\appendix

\twocolumn[
  \begin{@twocolumnfalse}
    \huge{\textbf{Appendix}}
    \vspace{5mm}
  \end{@twocolumnfalse}
]

\addcontentsline{toc}{section}{Appendices}
\renewcommand\thefigure{A\arabic{figure}}
\renewcommand{\thetable}{A\arabic{table}}
\setcounter{figure}{0}
\setcounter{table}{0}

\input{supp/_intro}

\input{supp/_analy_length}

\input{supp/_effect_hyperparam}

\input{supp/_more_qual}

\pagebreak

\clearpage
\input{figures/supfig3_qual}

\end{document}

%% file: __misc__.tex
\usepackage{graphicx}
\usepackage{amsmath}
\usepackage{amssymb}
\usepackage{booktabs}
\usepackage{times}
\usepackage{epsfig}
\usepackage{graphicx}
\usepackage{amsmath}
\usepackage{amssymb}
\usepackage{array}
\usepackage{csquotes}
\usepackage[accsupp]{axessibility} %

\usepackage[numbers,sort]{natbib}
\usepackage{xcolor, colortbl}
\usepackage{tabularx}
\usepackage{multirow}
\usepackage{enumitem}
\usepackage{bbm}
\usepackage{pifont}
\usepackage{capt-of}
\usepackage{array,multirow}
\usepackage{booktabs,siunitx,caption}
\usepackage[export]{adjustbox}

\newcolumntype{L}[1]{>{\raggedright\let\newline\\\arraybackslash\hspace{0pt}}m{#1}}
\newcolumntype{C}[1]{>{\centering\let\newline\\\arraybackslash\hspace{0pt}}m{#1}}
\newcolumntype{R}[1]{>{\raggedleft\let\newline\\\arraybackslash\hspace{0pt}}m{#1}}

\def\ie{\emph{i.e.}}
\def\eg{\emph{e.g.}}
\def\etal{\emph{et al.}}

\definecolor{Gray}{gray}{0.85}
\definecolor{brown}{rgb}{0.65, 0.16, 0.16}
\definecolor{purple}{rgb}{0.65, 0.1, 0.75}
\definecolor{yellow}{rgb}{0.75, 0.75, 0.0}
\definecolor{orange}{rgb}{1.0, 0.5, 0.2}
\definecolor{green}{rgb}{0, 0.8, 0.2}
\definecolor{red}{rgb}{1.0, 0.0, 0.0}
\definecolor{dblue}{rgb}{0.6, 0.1, 0.9}

\newcommand{\Fig}[1]{Fig.~\ref{fig:#1}}
\newcommand{\Sec}[1]{Sec.~\ref{sec:#1}}
\newcommand{\Eq}[1]{Eq.~(\ref{eq:#1})}
\newcommand{\Tbl}[1]{Table~\ref{tab:#1}}

\newcommand{\cmark}{\ding{51}}%
\newcommand{\xmark}{\ding{55}}%

\usepackage[breaklinks=true,letterpaper=true,colorlinks,bookmarks=true]{hyperref}

\usepackage[capitalize]{cleveref}
\crefname{section}{Sec.}{Secs.}
\Crefname{section}{Section}{Sections}
\Crefname{table}{Table}{Tables}
\crefname{table}{Tab.}{Tabs.}

%% file: _0_abstract.tex
\begin{abstract}
Referring image segmentation is an advanced semantic segmentation task where target is not a predefined class but is described in natural language.
Most of existing methods for this task rely heavily on convolutional neural networks, which however have trouble capturing long-range dependencies between entities in the language expression and are not flexible enough for modeling interactions between the two different modalities.
To address these issues, we present the first convolution-free model for referring image segmentation using transformers, dubbed ReSTR. 
Since it extracts features of both modalities through transformer encoders, it can capture long-range dependencies between entities within each modality.
Also, ReSTR fuses features of the two modalities by a self-attention encoder, which enables flexible and adaptive interactions between the two modalities in the fusion process.
The fused features are fed to a segmentation module, which works adaptively according to the image and language expression in hand. 
ReSTR is evaluated and compared with previous work on all public benchmarks, where it outperforms all existing models.

\end{abstract}

%% file: _1_introduction.tex
\vspace{-3mm}
\section{Introduction}
\label{sec:intro}
\vspace{-1mm}

Throughout the recent years, there have been witnessed remarkable advances in semantic segmentation in terms of both efficacy and efficiency~\cite{Fcn, Deeplabcrf, deeplab_v2, PSPNet, deconvnet, huang2019ccnet, zhao2020exploring}.
However, its application to real-world downstream tasks is still limited. 
Since the task is designed to deal with only a predefined set of classes (\eg, ``car'', ``person''),
semantic segmentation models are hard to address undefined classes and specific entities of user interest (\eg, ``a red Ferrari'', ``a man wearing a blue hat''). 

Referring image segmentation~\cite{hu2016segmentation} has been studied to resolve this limitation by segmenting an image region corresponding to a natural language expression given as query. 
As this task is no longer restricted by predefined classes, it enables a large variety of applications such as human-robot interaction and interactive photo editing. 
Referring image segmentation is however more challenging than semantic segmentation since it demands to comprehend individual entities and their relations expressed in the language expression (\eg, ``a car behind the taxi next to the building''), and to fully exploit such structured and relational information in the segmentation process.
For this reason, models for the task should be capable of capturing interactions between semantic entities in both modalities as well as joint reasoning over the two different modalities.
    
\input{figures/fig0_teaser}

Existing methods for referring image segmentation~\cite{hu2016segmentation, liu2017recurrent, li2018referring, margffoy2018dynamic, shi2018key, chen2019see, ye2019cross, hu2020bi, hui2020linguistic, huang2020referring, feng2021encoder} have adopted convolutional neural networks (CNNs) and recurrent neural networks (RNNs) to extract visual and linguistic features, respectively.
In general, these features are integrated into a multimodal feature map through convolution layers applied to a concatenation of the two features, so-called concatenation-convolution operation.
On top of the multimodal feature map, recent methods~\cite{ye2019cross, hu2020bi, hui2020linguistic, huang2020referring, feng2021encoder} further employ attention mechanisms~\cite{vaswani2017attention, wang2018non} so that the feature map effectively captures interactions between semantic entities.
The final multimodal features are then fed as input to a segmentation module.

Although these methods have shown remarkable results on the challenging task, they share the following limitations. 
First, they have trouble handling long-range interactions between semantic entities within each modality. 
Referring image segmentation requires to capture such interactions since language expressions often involve complicated relations between entities to precisely indicate target region.
In this aspect, both of CNNs and RNNs are limited due to the locality of their basic building blocks. 
Second, existing models have difficulty in modeling sophisticated interactions between the two modalities.
They aggregate visual and linguistic features through the concatenation-convolution operation, which is a fixed and handcrafted way of feature fusion and thus could not be sufficiently flexible and effective to handle a large variety of referring image segmentation scenarios.

To overcome the aforementioned limitations, we propose the first convolution-free model for Referring image Segmentation using TRansformers, dubbed ReSTR.
Its overall pipeline is illustrated briefly in~\Fig{teaser}.
First of all, ReSTR extracts visual and linguistic features through transformer encoders~\cite{vaswani2017attention}.
The two encoders, namely \emph{vision encoder} and \emph{language encoder}, take a set of non-overlapped image patches and that of word embeddings as input, respectively, and extract their features while considering their long-range interactions within each modality. 
By using transformers for both modalities, we take advantage of capturing global context from the beginning of feature extraction and unifying network topology for the two modalities~\cite{nagrani2021attention}.

Next, a self-attention encoder aggregates the visual and linguistic features into a patch-wise multimodal features. 
This multimodal fusion encoder enables sophisticated and flexible interactions between features of the two modalities thanks to its self-attention layers. 
Moreover, the fusion encoder takes a class seed embedding as another input.
The class seed embedding is transformed adaptively by the fusion encoder to a classifier for the target entity described in the language expression.

Finally, the outputs of the multimodal fusion encoder, \ie, the patch-wise multimodal features and the adaptive classifier, are fed as input to the segmentation decoder. 
The decoder computes the final segmentation map in a coarse-to-fine manner. 
The adaptive classifier is first applied to each multimodal feature as a classifier to examine whether each image patch contains a part of target entity. 
The coarse, patch-level prediction is then converted into a pixel-level segmentation map by a series of upsampling and linear layers. 
Thanks to the powerful transformer encoders, this simple and efficient decoder is able to produce accurate segmentation results, achieving the state of the art on four public benchmarks for referring image segmentation.

In summary, the contribution of this work is three-fold:
\begin{itemize}
    \vspace{-2mm}\item Our network is the first convolution-free architecture for referring image segmentation.
    It captures long-range interactions between vision and language modalities and unifies the network topology for the two different modalities by transformers.
    \vspace{-2.5mm}\item To encode the fine comprehension of the two modalities, we carefully design the multimodal fusion encoder with the class seed embedding which is transformed to an adaptive classifier for referring image segmentation.
    \vspace{-2.5mm}\item ReSTR achieves the state of the art on four public benchmarks without bells and whistles.
\end{itemize}

%% file: figures/fig0_teaser.tex
\begin{figure}[t!]
    \centering
    \includegraphics[width=0.96\linewidth]{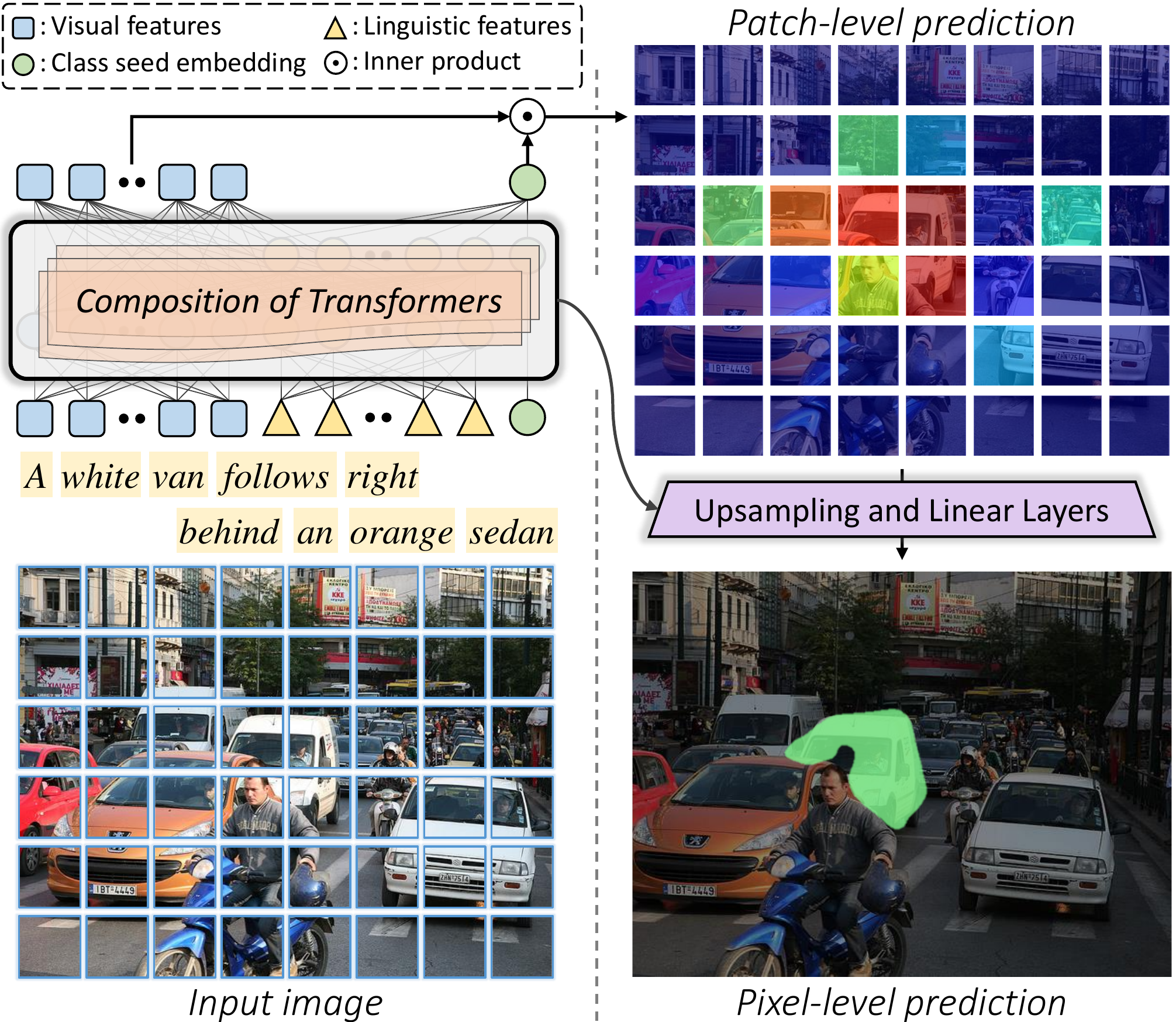}
\vspace{-1mm}
\caption{
Our convolution-free architecture for Referring image Segmentation using TRansformer (ReSTR) takes a set of non-overlapped image patches and that of word embeddings, and captures intra- and inter-modality interactions by transformers.
Then, ReSTR takes a class seed embedding to produce an adaptive classifier which examines whether each image patch contains a part of target entity.
Finally, a series of upsampling and linear layers computes a pixel-level prediction in a coarse-to-fine manner.
} \label{fig:teaser}
\vspace{-5mm}
\end{figure}

%% file: _2_relatedwork.tex
\vspace{-1mm}
\section{Related Work}
\label{sec:relatedwork}
\vspace{-1mm}
\subsection{Semantic Segmentation}
\vspace{-1.5mm}
Semantic segmentation has been significantly improved with the emergence of deep neural networks.
Based on a Fully Convolutional Network (FCN)~\cite{Fcn} for pixel-level prediction on an end-to-end framework, many approaches are proposed to overcome the several limitations of the network.
Since FCN predicts a coarse output mask, the early approaches~\cite{Crfrnn, randomwalk_net, Deeplabcrf, Liu15} focus on performing high-resolution predictions.
The former studies propose methods to extend the receptive field of CNN by dilated convolutions~\cite{deeplab_v2, DilatedConv} and to capture multiscale contexts by a feature pyramid pooling scheme~\cite{deeplab_v2, PSPNet, yang2018denseaspp}.
The several approaches propose encoder-decoder structures~\cite{deconvnet, Deconvnet2, lin2017refinenet, unet, deeplabv3} to model coarse-to-fine framework by multi-level feature fusion.  
Recently, semantic segmentation has been studied to capture contextual information by attention mechanism~\cite{zhao2018psanet, huang2019ccnet, zhang2019acfnet}.

However, the mentioned methods have used variants of FCN architecture that limit local context encoding by convolutional layers.
Moreover, since this task is defined to predict segmentation masks within only a predefined set of classes, semantic segmentation models have limitations to apply to real applications.

\input{figures/fig1_overall}

\vspace{-0.5mm}
\subsection{Referring Image Segmentation} 
\vspace{-1.5mm}
In contrast to predefined pixel-level classification as semantic segmentation, referring image segmentation aims at grouping the pixels as mask corresponded to a given natural language expression.
The pioneering work~\cite{hu2016segmentation} proposes extracting visual and linguistic features from CNN and RNN, respectively, and generating multimodal features by concatenating tiled linguistic features and visual feature maps.
Based on this framework, the early approaches suggest the methods to perform high-resolution prediction by ConvLSTM~\cite{liu2017recurrent} and the encoder-decoder architecture by intermediate connections~\cite{li2018referring}.
Follow-up studies propose an attention mechanism to fuse the visual and linguistic features and multi-level feature aggregation to produce high-resolution segmentation maps~\cite{chen2019see, ye2019cross, hu2020bi, feng2021encoder}.
Recent studies~\cite{kamath2021mdetr, ding2021vision} suggest a multimodal fusion encoder using transformers~\cite{vaswani2017attention} to capture long-range interactions between visual and linguistic features.

Unlike the existing work, we propose a new convolution-free architecture to encode contextual information at every stage of our model and efficiently transform a patch-level prediction to a high-resolution segmentation map in a coarse-to-fine manner.

\vspace{-0.5mm}
\subsection{Vision Transformer}
\vspace{-1.5mm}
From the introduction of transformers by~\cite{vaswani2017attention} as a self-attention module for NLP, many approaches adopt this module in computer vision tasks for the advantages of this module including long-range dependencies, dynamic kernel depended on input, and less visual inductive bias than CNNs.
Several studies employ transformers for an attention module in/on CNNs as a CNN-transformer hybrid network~\cite{wang2018non, ramachandran2019stand, zhao2020exploring, carion2020end, srinivas2021bottleneck, zheng2021rethinking, wang2021max}.
Recent approaches replace CNNs with transformers as a convolution-free architecture in image classification~\cite{dosovitskiy2020image, jaegle2021perceiver, liu2021swin, wang2021pyramid}, object detection~\cite{liu2021swin, wang2021pyramid}, semantic segmentation~\cite{strudel2021segmenter, liu2021swin, zheng2021rethinking, wang2021pyramid} and multimodal learning~\cite{nagrani2021attention}.
In particular, transformers are deployed to semantic segmentation tasks to overcome the inherent limitation of FCN-like architecture.
For example, Zheng~\etal~\cite{zheng2021rethinking} utilize transformer backbone as a global context feature extractor and then convolutional layers for a decoder in the hybrid manner.
Strudel~\etal~\cite{strudel2021segmenter} propose a convolution-free architecture for semantic segmentation by self-attention with visual features and a set of learnable classes embeddings.
Inspired by the paradigm, we adopt transformers for referring image segmentation for the above advantages and use an adaptive classifier as an extension of the learnable class queries used in semantic segmentation transformers~\cite{strudel2021segmenter, wang2021max}.

%% file: figures/fig1_overall.tex
\begin{figure*}[t!]
    \centering
    \includegraphics[width=0.98\linewidth]{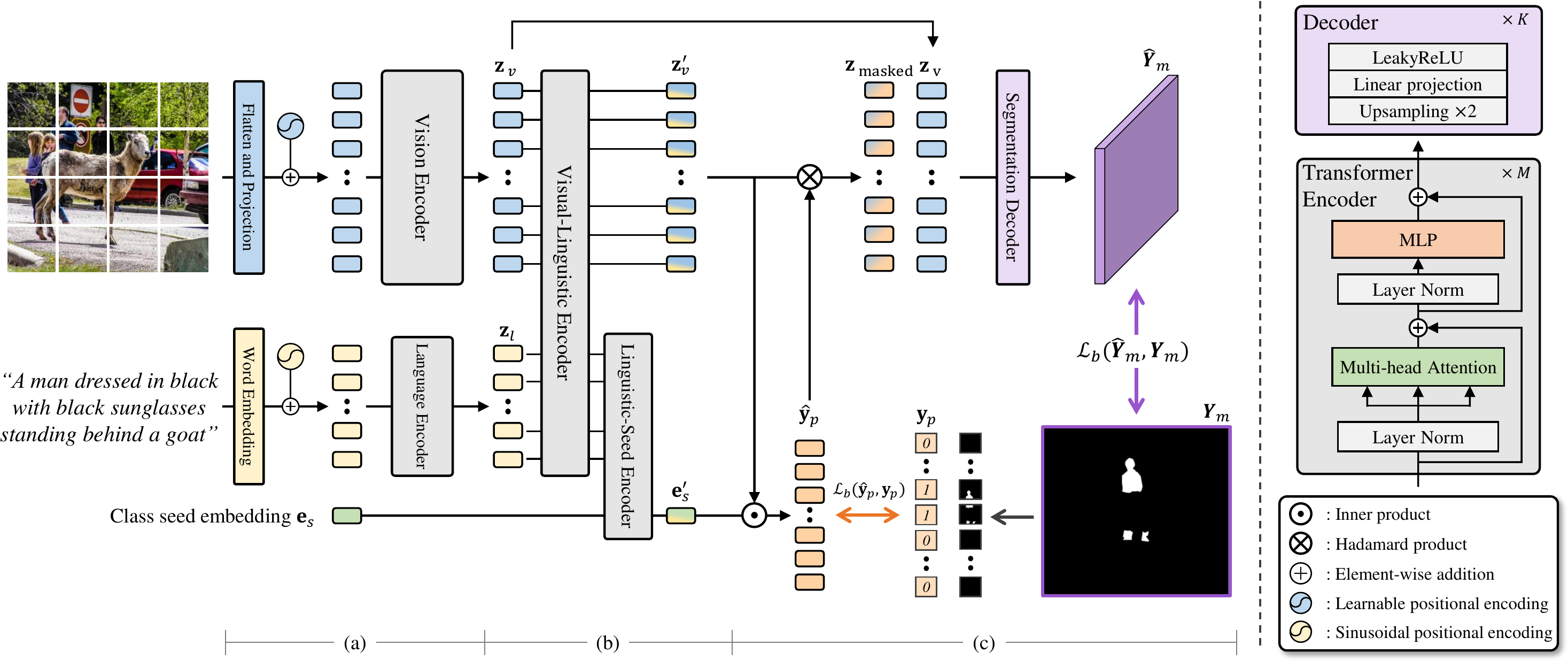}
\vspace{-1mm}
\caption{
(\textit{Left}) Overall architecture of ReSTR.
(a) The feature extractors for the two modalities are composed on transformer encoders, independently.
(b) The multimodal fusion encoder consists of the two transformer encoders: \textit{the visual-linguistic encoder} and \textit{the linguistic-seed encoder}.
(c) The coarse-to-fine segmentation decoder transforms a patch-level prediction to a pixel-level prediction.
(\textit{Right}) Transformer encoder used in all encoders and the composition of the coarse-to-fine segmentation decoder.
} \label{fig:overall}
\vspace{-5mm} 
\end{figure*}

%% file: _3_method.tex
\section{Proposed Method}
\vspace{-1mm}
This section elaborates on ReSTR, our convolutional-free transformer network for referring image segmentation.
Its detailed architecture is illustrated in~\Fig{overall}.
To capture long-range interactions for each modality, ReSTR first extracts visual and linguistic features by transformer encoders~\cite{vaswani2017attention} independently (\Sec{extraction}).
Then, it forwards visual and linguistic features in parallel to a multimodal fusion encoder to capture fine relations across these two modalities (\Sec{encoder}).
Finally, an efficient decoder for a coarse-to-fine segmentation converts patch-level prediction into high-resolution pixel-level prediction (\Sec{decoder}).

\subsection{Visual and Linguistic Feature Extraction}
\label{sec:extraction}
\vspace{-1mm}
To extract visual and linguistic features, we choose transformers~\cite{dosovitskiy2020image} for both modalities.
A transformer encoder is $M$ sequential transformers, each of which consists of Multi-headed Self-Attention (MSA), Layer Normalization (LN), and Multilayer Perceptron (MLP) blocks:
\begin{equation}
    \begin{split}
    \bar{\mathbf{z}}_{i+1} = \mathrm{MSA}(\mathrm{LN}(\mathbf{z}_{i})) + \mathbf{z}_{i},
    \end{split}
\end{equation}
\vspace{-5mm}
\begin{equation}
    \mathbf{z}_{i+1} = \mathrm{MLP}(\mathrm{LN}(\bar{\mathbf{z}}_{i+1})) + \bar{\mathbf{z}}_{i+1},
\end{equation}
where $\mathbf{z}_i \in\mathbb{R}^{N \times D}$ denotes input features of the $i$-th layer of the transformer encoder, $N$ is the input size of each modality, and $D$ is the channel dimension of the features.
LN is applied to the output of the transformer encoder.
MSA is composed of $k$ Self-Attention (SA) operations on queries $\mathbf{q}\in \mathbb{R}^{N \times D_{h}}$, keys $\mathbf{k}\in \mathbb{R}^{N \times D_{h}}$, and values $\mathbf{v}\in \mathbb{R}^{N \times D_{h}}$, which are obtained by linear projections of input features $\mathbf{z}$, independently:
\begin{equation}
    \mathrm{MSA}(\mathbf{z}) = [\mathrm{SA}_{1}(\mathbf{z}), \mathrm{SA}_{1}(\mathbf{z}),\cdot\cdot\cdot,\mathrm{SA}_{k}(\mathbf{z})]\mathbf{W}_{\mathrm{MSA}},
\end{equation}
\vspace{-5mm}
\begin{equation}
    \mathrm{SA}(\mathbf{z}) = A\mathbf{v},
\end{equation}
\vspace{-4mm}
\begin{equation}
    A = \mathrm{softmax}(\mathbf{q}\mathbf{k}^{\top} / \sqrt{D_{h}}),
\label{eq:atten}
\end{equation}
where $A\in\mathbb{R}^{N\times N}$ is dot-product attention, $[\cdot,\cdot]$ denotes concatenation, and $\mathbf{W}_{\mathrm{MSA}}\in \mathbb{R}^{kD_{h}\times D}$ is a linear projection.
$D_{h}$ is set to $D/k$ following~\cite{dosovitskiy2020image}.
The transformer encoder, which is composed of transformers, is denoted by $\mathrm{Transformers}(\cdot)$.

\noindent \textbf{Vision encoder.} An input image $\mathbf{x}^{v}\in \mathbb{R}^{H\times W\times C_{v}} $ is transformed to a set of patch embeddings $\mathbf{x}^{p}\in \mathbb{R}^{N_{v}\times D_{v}}$ by splitting the input image into the patches without overlapping and mapping them with a linear projection.
Let $N_{v}=HW/P^{2}$ be the number of patches, $P$ be the patch size, and $D_{v}$ be the projected channel dimension.
We add learnable 1D positional encoding $\mathbf{E}^{v}_{\mathrm{pos}}\in \mathbb{R}^{N_{v}\times D_{v}}$ to the patch embeddings to obtain input to the vision encoder, $\mathbf{z}^{v}_{0}=\mathbf{x}^{p}+\mathbf{E}^{v}_{\mathrm{pos}}$.
We feed $\mathbf{z}^{v}_{0}$ into the vision encoder to produce the patch-wise visual features $\mathbf{z}_{v}\in\mathbb{R}^{N_{v}\times D_{v}}$:
\begin{equation}
    \mathbf{z}_{v}= \mathrm{Transformers}(\mathbf{z}^{v}_{0};\boldsymbol{\theta}_{v}),
\end{equation}
where $\boldsymbol{\theta}_{v}$ are the parameters of the vision encoder.

\noindent \textbf{Language encoder.} We transform a natural language expression to a set of word embeddings $\mathbf{x}^{l}\in\mathbb{R}^{N_{l}\times C_{l}}$, where $N_{l}$ is the maximum length of sentence and $C_{l}$ is the dimension of the word embeddings.
We add a sinusoidal 1D positional encoding~$\mathbf{e}^{l}_{\mathrm{pos}} \in \mathbb{R}^{N_{l}\times C_{l}}$ to the word embeddings, as $\mathbf{z}^{l}_{0}=\mathbf{x}^{l}+\mathbf{e}^{l}_{\mathrm{pos}}$.
Linguistic features $\mathbf{z}_{l}\in\mathbb{R}^{N_{l}\times D_{l}}$ are generated by feeding $\mathbf{z}^{l}_{0}$ into the language encoder which consists of transformers.

\subsection{Multimodal Fusion Encoder}
\label{sec:encoder}
\vspace{-1mm}
The multimodal fusion encoder consists of two transformer encoders, namely \textit{visual-linguistic encoder} and \textit{linguistic-seed encoder} as shown in~\Fig{overall}~(b). 
In detail, we use the visual features $\mathbf{z}_{v}$, the linguistic features $\mathbf{z}_{l}$ and a class seed embedding $\mathbf{e}_{s}\in \mathbb{R}^{1\times D}$ as input for the multimodal fusion encoder.
$\mathbf{e}_{s}$ is the trainable parameters, initialized randomly.
We first normalize $\mathbf{z}_{v}$ and $\mathbf{z}_{l}$ and feed each of them into a different linear layer to adjust their channel dimension to be the same as $D$.
Then, the visual-linguistic encoder takes the visual and linguistic features as inputs to produce patch-wise multimodal features $\mathbf{z}_{v}^{\prime}\in \mathbb{R}^{N_v\times D}$:
\begin{equation}
    [\mathbf{z}_{v}^{\prime}, \mathbf{z}_{l}^{\prime}] = \mathrm{Transformers}([\mathbf{z}_{v}, \mathbf{z}_{l}];\boldsymbol{\theta}_{vl}),
\end{equation}
where $\boldsymbol{\theta}_{vl}$ are the parameters of the visual-linguistic encoder, $\mathbf{z}_{l}^{\prime} \in \mathbb{R}^{N_l \times D}$ denotes visual-attended linguistic features.
Since the visual and linguistic features are fed into the visual-linguistic encoder in parallel, we obtain the patch-wise multimodal features by fine and flexible interactions between the visual and linguistic features.

Then, we feed the class seed embedding $\mathbf{e}_{s}$ and the visual-attended linguistic features $\mathbf{z}_{l}^{\prime}$ into the linguistic-seed encoder:
\begin{equation}
    \mathbf{e}_{s}^{\prime} = \mathrm{Transformers}([\mathbf{z}_{l}^{\prime}, \mathbf{e}_{s}];\boldsymbol{\theta}_{ls}),
\end{equation}
where $\boldsymbol{\theta}_{ls}$ are the parameters of transformers for the linguistic-seed encoder, and $\mathbf{e}_{s}^{\prime} \in \mathbb{R}^{1\times D}$ is an adaptive classifier.
Since a single fixed classifier is not sufficient for referring segmentation where a target mask varies by a language expression in hand, $\mathbf{e}_{s}^{\prime}$ acts as an adaptive classifier that examines if each patch contains a part of a target entity.

The multimodal fusion encoder is designed to produce the adaptive classifier that satisfies the following two requirements demanded in referring image segmentation.
First, since referring image segmentation aims to segment a region corresponding to a language expression, the adaptive classifier should comprehend fine relations of the language expression.
Moreover, since an input image has regions irrelevant to the language expression (\eg, background), the class seed embedding directly attending to the visual information can lead to an adaptive classifier corrupted by the irrelevant regions.
Nevertheless, since the appearance of the target entity described in a language expression can differ by images, it is beneficial to produce the adaptive classifier using the visual-attended linguistic features.

Therefore, we build the multimodal fusion encoder using these two transformer encoders alternatively to generate the adaptive classifier that meets the aforementioned conditions.
We empirically verify the superiority of our multimodal fusion encoder in ~\Sec{variant}.

\subsection{Coarse-to-Fine Segmentation Decoder}
\label{sec:decoder}
\vspace{-1mm}
A patch-level prediction $\mathbf{\hat{y}}_{p}\in \mathbb{R}^{N_{v}\times 1}$ is calculated by an inner product between the patch-wise multimodal features $\mathbf{z}_{v}^{\prime}$ and the adaptive classifier $\mathbf{e}_{s}^{\prime}$:
\begin{equation}
    \mathbf{\hat{y}}_{p} = \sigma\bigg(\frac{\mathbf{z}_{v}^{\prime}{\mathbf{e}_{s}^{\prime\top}}}{\sqrt{D}}\bigg),
\label{eq:p_patch}
\end{equation}
where $\sigma$ is the sigmoid function and $\sqrt{D}$ is a normalization factor~\cite{vaswani2017attention}.

We suggest an efficient segmentation decoder to compensate for the low-resolution patch-level prediction (\eg, $N_v={H}/{P}\times {W}/{P}$).
First, the decoder produces masked multimodal features $\mathbf{z}_{\mathrm{masked}}\in \mathbb{R}^{N_{v}\times D}$:
\begin{equation}
    \mathbf{z}_{\mathrm{masked}} = \mathbf{z}_{v}^{\prime} \otimes \mathbf{\hat{y}}_{p},
\end{equation}
where $\otimes$ denotes Hadamard product operation over the channel dimension $D$.
Then, before forwarding to the segmentation decoder, we concatenate the patch-wise visual features and the masked multimodal features as $[\mathbf{z}_{v}, \mathbf{z}_\mathrm{masked}]\in \mathbb{R}^{N_{v}\times 2D}$ to guide the segmentation decoder through visual semantics.
The segmentation decoder is composed of $K$ sequential blocks, each of which consists of upsampling with factor 2, linear projection with channel reduction by $1/2$ of input dimension, and activation function, where $K=\log{P}$ and $P$ is the patch size.
Finally, the output features of the decoder is fed into a linear projection and reshaped to generate a pixel-level prediction $\hat{Y}_{m}\in \mathbb{R}^{H\times W\times 1}$.
At inference time, we only use the pixel-level prediction $\hat{Y}_{m}$ as the final prediction. 

\label{sec:training}
For patch-level classification, we generate patch-level labels by splitting the ground-truth label $Y_{m}\in \mathbb{R}^{H \times W \times 1}$ into a set of patch labels whose number of patches is the same as the patch-level prediction $\mathbf{\hat{y}}_{p}\in \mathbb{R}^{N_v \times 1}$ by following criteria:
\begin{equation} 
    \mathbf{y}^{i}_{p} = 
    \begin{cases}
            1,  & \textrm{ if } h(p_{ij}) > \tau \\
            0, & \textrm{ otherwise}
    \end{cases},
\label{eq:patchlabel}
\end{equation}
where $\mathbf{y}^{i}_{p}$ denotes the patch-level labels of the $i$-th patch $p_i$, $j$ is the number of pixels in a patch, $h(\cdot)$ indicates the average pooling over spatial dimension, and $\tau$ is a thresholding hyperparameter.

The network is trained by the binary cross-entropy loss $\mathcal{L}_{b}(\hat{Y}, Y)$ on the patch-level prediction $\mathbf{\hat{y}}_{p}$ and the pixel-level prediction $ \hat{Y}_{m}$:
\begin{equation}
    \mathcal{L}(\mathbf{\hat{y}}_{p}, \mathbf{y}_{p}, \hat{Y}_{m}, Y_{m}) = \lambda \mathcal{L}_{b}(\mathbf{\hat{y}}_{p}, \mathbf{y}_{p}) + \mathcal{L}_{b}(\hat{Y}_{m}, Y_{m}), \label{eq:tot_loss}
\end{equation}
where $\lambda$ is a balancing hyperparameter.

%% file: _4_experiments.tex
\input{tables/comp_sota}

\section{Experiments}
\label{sec:experiment}
\vspace{-1mm}
\subsection{Experimental Setting}
\vspace{-1mm} \noindent \textbf{Datasets.} 
We conduct experiments on four datasets, ReferIt~\cite{kazemzadeh2014referitgame}, UNC~\cite{yu2016modeling}, UNC+~\cite{yu2016modeling}, and Gref~\cite{mao2016generation}, which are widely used in referring image segmentation task.
ReferIt~\cite{kazemzadeh2014referitgame} contains 19,894 images with 130,525 language expressions for 96,654 masks which are collected from IAPR TC-12~\cite{escalante2010segmented}.
{UNC}, {UNC+}, and {Gref} are collected from COCO~\cite{Mscoco} dataset.
UNC and UNC+ consist of 19,994 images with 142,209 language expressions for 50,000 masks and 19,992 images with 141,564 language expressions for 49,856 masks, respectively.
The difference between UNC and UNC+ is that UNC+ does not contain the words that indicate location properties (\eg, left, top, front) in expressions and contains the only appearance expressions.
Gref contains 25,711 images with 104,560 language expressions for 54,822 objects.

\noindent \textbf{Implementation details.}
We use ViT-B-16~\cite{dosovitskiy2020image} pretrained on ImageNet-21K~\cite{Imagenet} for the vision encoder which has 12 layers, 16 patch size, 768 channel dimensions, 12 heads of MSA, and 3,072 dimensions of channel expansion in MLP.
We use pretrained GloVe~\cite{pennington2014glove} embeddings for language expressions.
The language encoder consists of 6 transformer layers, and has 300 channel dimensions as GloVe embeddings, 12 heads of MSA and 3,072 dimensions of channel expansion in MLP.
The maximum length of a language expression $N_l$ is set to 20 following previous work.
The multimodal fusion encoder consists of the same transformer as the vision encoder.
The number of layers of the segmentation decoder is 4 since the patch size is 16. 
In all experiments, the models are optimized by AdamW~\cite{loshchilov2017decoupled} with weight decay of $5e-4$; the initial learning rate is $1e-5$ and decreases with polynomial decay~\cite{Deeplabcrf}.
We set a batch size of 8 and train for 400,000 iterations with warm-up period for 40,000 iterations to reach the initial learning rate.
We resize input images to $480\times480$.
We set $\tau$ in~\Eq{patchlabel} and $\lambda$ in~\Eq{tot_loss} to 0.8 and 0.1 for all experiments, respectively.

\noindent \textbf{Evaluation protocol.} Following previous work~\cite{hu2016segmentation, liu2017recurrent}, we adopt the cumulative Intersection-over-Union (IoU) metric, where total intersections are divided by the total unions over all test samples.
Then, we evaluate the accuracy at the $\{0.5, 0.6, 0.7, 0.8, 0.9\}$ IoU thresholds.

\input{tables/word_leng_analy}

\subsection{Comparisons with the State of the Art}
\label{sec:sota}
We compare ReSTR with other referring image segmentation models on four benchmarks.
As summarized in~\Tbl{comp_sota}, ReSTR achieves outstanding performance without inefficient postprocessing (\eg,~DenseCRF~\cite{Fullycrf}) compared with the previous arts on all public benchmarks except for UNC+ \textit{testB} set.
Following~\cite{liu2017recurrent}, we discuss the relationship between language expression length and performance as summarized in~\Tbl{lang_vs_iou}.
The results demonstrate the ReSTR clearly outperforms previous methods on most groups of expression length except for the 1-5 length group on Gref \textit{val} set.
Moreover, the performance of ACM using an attention mechanism for long-range interactions between the two modalities drops 13.71\%p from 1-5 to 11-20 length group on the Gref \textit{val} set, while that of ReSTR drops by 6.81\%p.
It demonstrates that our method is better to capture the long-range interactions between the two modalities compared to previous methods.
Note that the recent methods~\cite{yang2021bottom, jing2021locate, ding2021vision} use a visual backbone pretrained on COCO object detection dataset and evaluate their models on only three benchmarks based on COCO dataset.
In contrast, our visual backbone is pretrained for ImageNet classification, and ReSTR is evaluated on all benchmarks.

\input{tables/variant_encoder}

\subsection{Analysis of Variants of Fusion Encoder}
\label{sec:variant}
To verify our design choice for the multimodal fusion encoder, we investigate variants of the fusion encoder. 
We use 4 transformer layers, denoted as $\{f_1, f_2, f_3, f_4\}$, in all variants of the encoder.

First, as illustrated in~\Fig{var}(a), we present a variant of the fusion encoder which takes all features simultaneously as inputs, denoted as Vanilla Multimodal Encoder (VME).
Since all inputs are given in parallel, VME can learn the fine relations between all features.
However, the adaptive classifier can be undesirably biased to the visual features by the imbalance of the length of features between visual and linguistic features ($N_{v}\gg N_{l}$).
As shown in~\Tbl{ablencoder}(a), we measure attention scores of the visual and linguistic features to the class seed embedding.
In detail, we split the attentions of the class seed embedding $\mathbf{a} \in \mathbb{R}^{1\times (N_v+N_l+1)}$ in the attention matrix $A$ in~\Eq{atten} into the visual and linguistic attentions $\mathbf{a}_v \in \mathbb{R}^{1\times N_v}$ and $\mathbf{a}_l \in \mathbb{R}^{1\times N_l}$, respectively.
Then, we sum each modality attention across feature dimension to obtain $a^i_v$ and $a^i_l$ as the attention score of VME of $i$-th transformer layer.
Finally, we average the attention score of each layer over the dataset.
The results demonstrate that the attentions for the class seed embedding are biased to the visual features.
We hypothesize that the bias of attentions results from the imbalance of the length of features between the visual and linguistic features, which is $N_v:N_l=900:20$ in our experiments, that leads to the adaptive classifier capturing less fine relations of a language expression.

\input{figures/fig2_var}
\input{tables/ablation}

To resolve this problem, we consider disconnecting interactions between the visual features and the class seed embedding as illustrated in~\Fig{var}(b), denoted as Independent Multimodal Encoder (IME).
In other words, the class seed embedding interacts with only the linguistic features.
Therefore, IME restricts the class seed embedding from being adaptively transformed to an adaptive classifier with the visual information.

To this end, we propose a structure that indirectly conjugates the class seed embedding and the visual features with the linguistic features as medium, denoted as indirect Conjugating Multimodal Encoder (CME) as illustrated in~\Fig{var}(c).
As mentioned in~\Sec{encoder}, the design aims to avoid interaction between the irrelevant visual features and the class seed embedding by indirectly interactions via the linguistic features.
Furthermore, CME produces the adaptive classifier for the target entity described in the language expression by fine interactions between the linguistic features and the class seed embedding.

As summarized in~\Tbl{ablencoder}(b), we compare the three variants of the multimodal fusion encoder on performance, computational cost (MACs), and the number for parameters (\# params) of these encoders without the segmentation decoder.
These results demonstrate the superiority of CME over the other variants of the fusion encoder in performance and efficiency.
In addition, we also experiment CME with weight sharing (CME$^{\dagger}$) between transformer layers of the visual-linguistic encoder and between those of the linguistic-seed encoder.
The result shows CME$^{\dagger}$ is still better performance with lower parameters and computational cost than the other variants.

\input{figures/fig4_qual}

\subsection{In-depth Analysis of ReSTR}
\vspace{-1mm}
\label{sec:ablation}

We investigate our framework on the \textit{val} set of Gref dataset which contains the longer and more complicated language expressions than the others.

\vspace{1mm} \noindent \textbf{Effect of the number of transformer layers in the multimodal fusion encoder.}
We study the impact of the number of transformer layers in the multimodal fusion encoder by varying the number of transformers to \{2, 4, 6\}.
Since the multimodal fusion encoder is composed of two transformer encoders, the encoder always has an even number of the transformer layers.
As summarized in~\Tbl{ablation}, the performance is fairly increased until using 4 transformer layers and marginally increased using 6 transformer layers.

\vspace{1mm} \noindent \textbf{Effect of the segmentation decoder.}
We investigate the contribution of the segmentation decoder.
As summarized in~\Tbl{ablation}, the decoder improves IoU by 1.67\%p when used along with the 4 transformer layers in the fusion encoder.
However, when with the fusion encoder with 2 transformer layers, the improvement made by the segmentation decoder is only 0.31\%p.
When coupled with the shallow fusion encoder that produces relatively larger potion of false patch-level predictions, the effect of the segmentation decoder is marginal since it is trained to refine the mask of the positive patches.
The results demonstrate that the decoder is specialized to refine a patch-level prediction to a fine pixel-level prediction.
Note that the analysis of the segmentation decoder is examined except for the fusion encoder with 6 transformer layers due to the memory shortage.

\input{tables/computation}

\vspace{1mm} \noindent \textbf{Effect of weight sharing.}
In~\Tbl{ablation}, we also present the performance of the model with weight sharing.
Using weight sharing, the number of parameters remains the same regardless of the number of transformer layers that the multimodal fusion encoder contains.
The results show that the performance degradation incurred by weight sharing is marginal.
It demonstrates that ReSTR could be used in an efficient manner with little loss of performance using weight sharing.

\vspace{1mm} \noindent \textbf{Qualitative analysis.}
As illustrated on \Fig{qualitative}, the patch-level predictions of ReSTR are roughly localized on the target patches and the boundaries of relational objects.
Then, the patch-level predictions are transformed to fine pixel-level predictions by the segmentation decoder in a coarse-to-fine manner.
Moreover, in~\Fig{varisent}, we provide the visualization examples of the predictions when varying language expressions are given as queries.
These visualizations show that ReSTR is able to predict the segmentation masks corresponding to different language expressions on an image.

\vspace{1mm} \noindent \textbf{Computation cost analysis.}
In~\Tbl{computation}, we present the number of parameters and MACs of ReSTR and recent studies whose codes are publicly available.
ReSTR achieves the best accuracy with the least computation since it employs the efficient segmentation decoder.
Also, the size of the visual feature used in previous work is 4 times bigger than ours.

%% file: tables/comp_sota.tex
\begin{table*}[t!]
\setlength{\tabcolsep}{4pt}
\centering

\scalebox{0.96}{
\begin{tabular}{p{2.5cm}<{}|p{1cm}<{\centering}|p{1.1cm}<{\centering}|p{1.1cm}<{\centering}p{1.1cm}<{\centering}p{1.1cm}<{\centering}|p{1.1cm}<{\centering}p{1.1cm}<{\centering}p{1.1cm}<{\centering}|p{1.1cm}<{\centering}}
\toprule
\multirow{2}{*}{Methods} & \multirow{2}{*}{DCRF}
&\multicolumn{1}{c|}{ReferIt} & \multicolumn{3}{c|}{UNC}
&\multicolumn{3}{c|}{UNC+} &\multicolumn{1}{c}{Gref}\\
                                    &       & \textit{test}    & \textit{val} & \textit{testA} & \textit{testB}   & \textit{val} & \textit{testA} & \textit{testB}  & \textit{val}  \\ \midrule
LSTM-CNN~\cite{hu2016segmentation}  &       &48.03 &-     &-     &-     &-     &-     &-     &28.14\\
RMI~\cite{liu2017recurrent}         &\cmark &58.73 &45.18 &45.69 &45.57 &29.86 &30.48 &29.50 &34.52\\
DMN~\cite{margffoy2018dynamic}      &       &52.81 &49.78 &54.83 &45.13 &38.88 &44.22 &32.29 &36.76\\
RRN~\cite{li2018referring}          &\cmark &63.63 &55.33 &57.26 &53.95 &39.75 &42.15 &36.11 &36.45\\
CMSA~\cite{ye2019cross}             &\cmark &63.80 &58.32 &60.61 &55.09 &43.76 &47.60 &37.89 &39.98\\
STEP~\cite{chen2019see}             &       &64.13 &60.04 &63.46 &57.97 &48.19 &52.33 &40.41 &46.40\\
BRINet~\cite{hu2020bi}              &\cmark &63.46 &61.35 &63.37 &59.57 &48.57 &52.87 & 42.13 &48.04\\
LSCM~\cite{hui2020linguistic}       &\cmark &66.57 &61.47 &64.99 &59.55 &49.34 &53.12 & 43.50 &48.05\\
CMPC~\cite{huang2020referring}      &\cmark &65.53 &61.36 &64.54 &59.64 &49.56 &53.44 &43.23 &49.05\\
ACM~\cite{feng2021encoder}          &       &66.70 & 62.76 & 65.69 &59.67 & 51.50 & 55.24 &43.01 & 51.93 \\
BUSNet~\cite{yang2021bottom}        &\cmark &- & 63.27 & 66.41 & 61.39 & 51.76 & 56.87 & 44.13 &50.56 \\ 
LTS~\cite{jing2021locate}           &       &- & 65.43 & 67.76 & \underline{63.08} & 54.21 & 58.32 & 48.02 &\underline{54.40} \\ 
VLT~\cite{ding2021vision}           &       &- & \underline{65.65} & \underline{68.29} & 62.73 & \underline{55.50} & \underline{59.20} &\textbf{49.36} & 52.99 \\ \midrule
ReSTR (Ours)                          &   & \textbf{70.18} & \textbf{67.22} & \textbf{69.30} & \textbf{64.45}  & \textbf{55.78} & \textbf{60.44} & \underline{48.27} & \textbf{54.48} \\
\bottomrule
\end{tabular}
}
\vspace{-1mm}
\caption{
Quantitative results on four datasets in IoU (\%). 
DCRF denotes using post-procession by DenseCRF~\cite{Fullycrf}.
The best results are in bold, while second-best ones are underlined.}
\vspace{-5mm}
\label{tab:comp_sota}
\end{table*}

%% file: tables/word_leng_analy.tex
\begin{table}[t!]
\centering
\small
\label{tab:my-table}
\scalebox{0.92}{
\begin{tabular}{l|c|cccc}
\toprule
                         & Length & 1-5 & 6-7 & 8-10 & 11-20 \\ \midrule
\multirow{4}{*}{Gref}    & R+RMI~\cite{liu2017recurrent}  & 35.34 &31.76 &30.66 &30.56     \\
                         & BRINet~\cite{hu2020bi} & 51.93 & 47.55 & 46.33 & \underline{46.49} \\
                         & ACM~\cite{feng2021encoder} & \textbf{59.92} & \underline{52.94} & \underline{49.56} & 46.21 \\
                         & ReSTR (Ours)   &  \underline{58.72}   &  \textbf{53.47}   &  \textbf{53.96} & \textbf{51.91}      \\ \midrule\midrule
                         & Length & 1-2 & 3   & 4-5  & 6-20  \\ \midrule
\multirow{4}{*}{UNC}     & R+RMI~\cite{liu2017recurrent}  & 44.51 & 41.86 & 35.05 & 25.95  \\
                         & BRINet~\cite{hu2020bi} & 65.99 & 64.83 & 56.97 & 45.65  \\
                         & ACM~\cite{feng2021encoder} & \underline{68.73} &\underline{65.58} & \underline{57.32} & \underline{45.90} \\
                         & ReSTR (Ours)   & \textbf{72.38}    &  \textbf{69.46}  & \textbf{61.19}  & \textbf{50.21}  \\ \midrule\midrule
                         & Length & 1-2 & 3   & 4-5  & 6-20  \\ \midrule
\multirow{4}{*}{UNC+}    & R+RMI~\cite{liu2017recurrent}  & 35.72 & 25.41 &  21.73 & 14.37 \\
                         & BRINet~\cite{hu2020bi} & 59.12 & 46.89 & 40.57 & 31.32  \\
                         & ACM~\cite{feng2021encoder} & \underline{61.62} & \underline{52.18} & \underline{43.46} & \underline{31.52} \\
                         & ReSTR (Ours)   &   \textbf{65.72}  &  \textbf{54.81}   &  \textbf{47.65}    &   \textbf{37.02}    \\ \midrule\midrule
                         & Length & 1   & 2   & 3-4  & 5-20  \\ \midrule
\multirow{4}{*}{ReferIt} & R+RMI~\cite{liu2017recurrent}  & 68.11 & 52.73 & 45.69 & 34.53  \\
                         & BRINet~\cite{hu2020bi} & 75.28 & 62.62 & 56.14 & 44.40 \\
                         & ACM~\cite{feng2021encoder} & \underline{78.19} & \underline{66.63} & \underline{60.30} & \underline{46.18}  \\
                         & ReSTR (Ours)   & \textbf{80.82}   &   \textbf{69.78}  &  \textbf{63.66}    &   \textbf{50.73}  \\ \bottomrule
\end{tabular}
}
\vspace{-1mm}
\caption{
Performance according to variants of referring length on Gref, UNC, UNC+ and ReferIt in IoU (\%).
The best results are in bold, while second-best ones are underlined.
}
\vspace{-5mm}
\label{tab:lang_vs_iou}
\end{table}

%% file: tables/variant_encoder.tex
\begin{table}[t!]
    \begin{subtable}[h]{0.17\textwidth}
        \centering
        \scalebox{0.84}{
        \begin{tabular}{lcc}
        \toprule
        Layer & $a_v$ & $a_l$                        \\ \midrule
        $f_1$ & 82.4 & 16.8                        \\
        $f_2$ & 98.9 & 1.0                         \\
        $f_3$ & 98.7 & 1.2                         \\
        $f_4$ & 98.1   & 1.7                        \\
        \bottomrule
        \end{tabular}
        }
        \caption{}
        \label{tab:attn_a}
     \end{subtable}
    \hfill
    \begin{subtable}[h]{0.30\textwidth}
        \centering
        \scalebox{0.84}{
        \begin{tabular}{lccc}
        \toprule
        \multicolumn{1}{l}{Encoder} & \# params & MACs     & IoU      \\ \midrule
        VME                         & 28.35M    & 31.36G    & 51.27    \\
        IME                         & 28.35M    & 15.96G     &  45.89     \\
        CME                         & 28.35M    & 15.96G     & \textbf{52.81}      \\
        CME$^{\dagger}$             & 14.18M    & 15.96G     & \underline{52.79}     \\
        \bottomrule
        \end{tabular}
       }
        \caption{}
       \label{tab:variantencoder}
    \end{subtable}
    \vspace{-1mm}
    \caption{
    (a) Averaged attention score (\%) of the visual and linguistic features to the class seed embedding at each transformer layer of VME on Gref \textit{train} set.
    (b) Performance of the variants of the multimodal fusion encoder on Gref \textit{val} set in IoU (\%). 
    $\dagger$ denotes the fusion encoder with weight sharing.
    The best results are in bold, while second-best ones are underlined.}
    \vspace{-5mm}
     \label{tab:ablencoder}
\end{table}

%% file: figures/fig2_var.tex
\begin{figure*}[t!]
    \centering
    \includegraphics[width=0.95\linewidth]{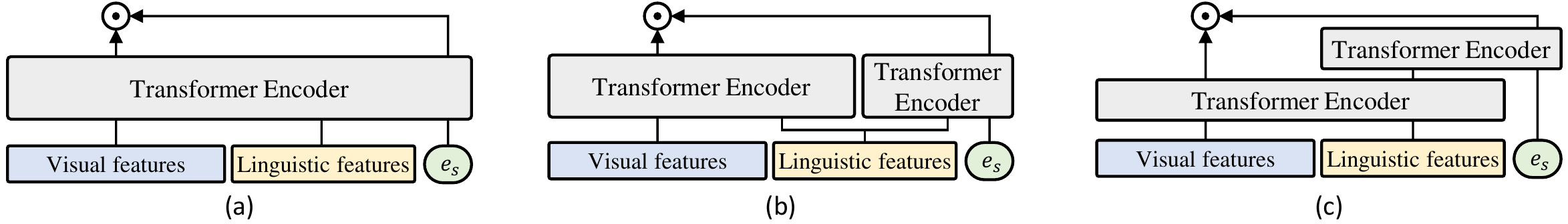}
\vspace{-1mm}
\caption{
The variants of the multimodal fusion encoder based on transformer architecture.
(a) Self-attention fusion encoder on all sequences in parallel.
(b) Independent fusion encoder between the visual features and the class seed embedding.
(c) Indirect conjugating fusion encoder between the visual features and the class seed embedding.
} \label{fig:var}
\vspace{-2mm}
\end{figure*}

%% file: tables/ablation.tex
\begin{table*}[t!]
\begin{center}
\scalebox{0.94}{
\begin{tabular}{p{1.2cm}<{\centering}|p{1.1cm}<{\centering}|p{1.2cm}<{\centering}|p{1.4cm}<{\centering}p{1.4cm}<{\centering}p{1.4cm}<{\centering}p{1.4cm}<{\centering}p{1.4cm}<{\centering}|p{1.4cm}<{\centering}}
\toprule
\multicolumn{2}{c|}{Encoder} & \multirow{2}{*}{Decoder} & \multirow{2}{*}{Prec@0.5}  &  \multirow{2}{*}{Prec@0.6}  & \multirow{2}{*}{Prec@0.7}  & \multirow{2}{*}{Prec@0.8}  & \multirow{2}{*}{Prec@0.9} &\multirow{2}{*}{IoU}  \\ 
\multicolumn{2}{c|}{\textit{\# layers}~~~~~~\textit{w share}} &  &  & &  &  &  & \\ \midrule
\multirow{2}{*}{2}  &        &        & 52.60 & 45.59 & 36.59 & 23.54 & 5.23 & 48.12 \\
                    &        & \cmark & 52.86 & 36.61 & 38.93 & 26.37 & 7.90 & 48.43 \\ \midrule
\multirow{3}{*}{4}  &        &        & 61.77  & 55.86  & 46.86  & 30.88  & 8.18  & 52.81 \\
                    &        & \cmark & 64.91  & 59.94  & 51.73  & 37.70  & 12.23 & \textbf{54.48} \\
                    & \cmark & \cmark & 64.27  & 59.01  & 50.70  & 35.85  & 11.46 & 54.07 \\ \midrule
\multirow{2}{*}{6}  &        &        & 63.36  & 57.88  & 48.75  & 33.46  & 8.75  & 52.84 \\
                    & \cmark &        & 63.05 & 57.32 & 48.19 & 32.47 & 8.47 & 52.59  \\
\bottomrule
\end{tabular}
}
\end{center}
\vspace{-4mm}
\caption{Performance for ablation study of ReSTR on Gref \textit{val} set. 
\textit{\# layers} denotes the number of transformer layers in the multimodal fusion encoder.
\textit{w share} denotes weight sharing of the multimodal fusion encoder.}
\vspace{-3mm}
\label{tab:ablation}
\end{table*}

%% file: figures/fig4_qual.tex
\begin{figure*}[t!]
\centering
\includegraphics[width=0.98\linewidth]{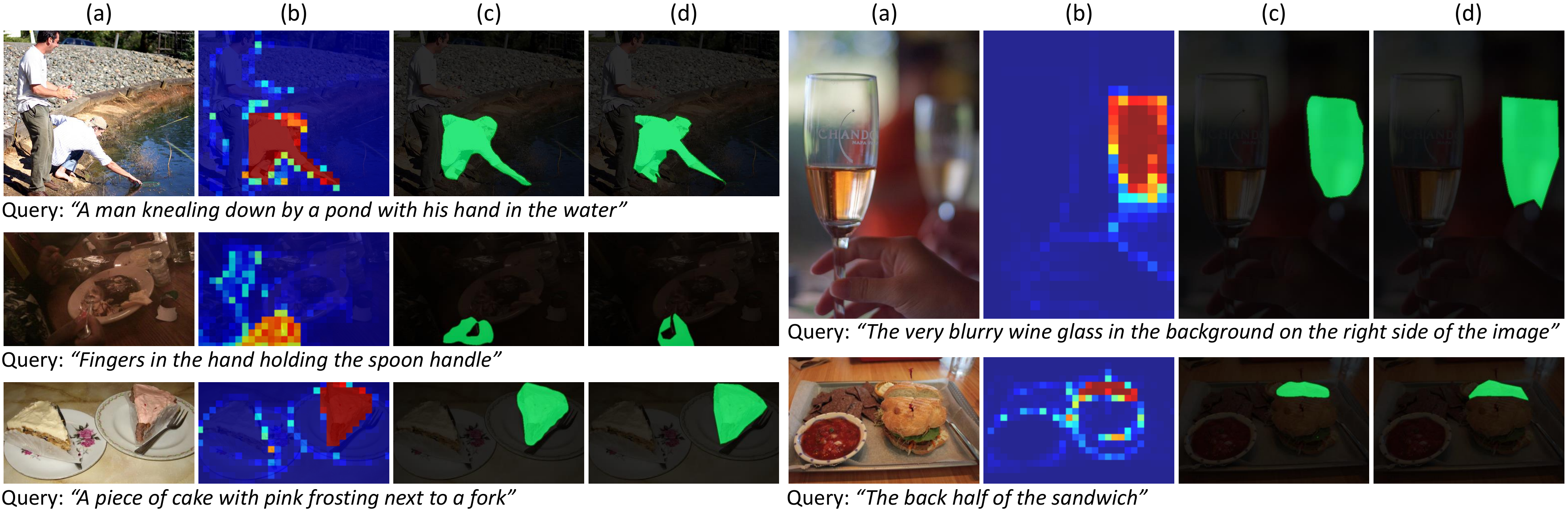}
\vspace{-1mm}
\caption{
Qualitative results of ReSTR on Gref \textit{val} set.
(a) Input image.
(b) Patch-level prediction.
(c) ReSTR.
(d) Ground truth.
} \label{fig:qualitative}
\end{figure*}

\begin{figure*}[t!]
\centering
\includegraphics[width=\linewidth]{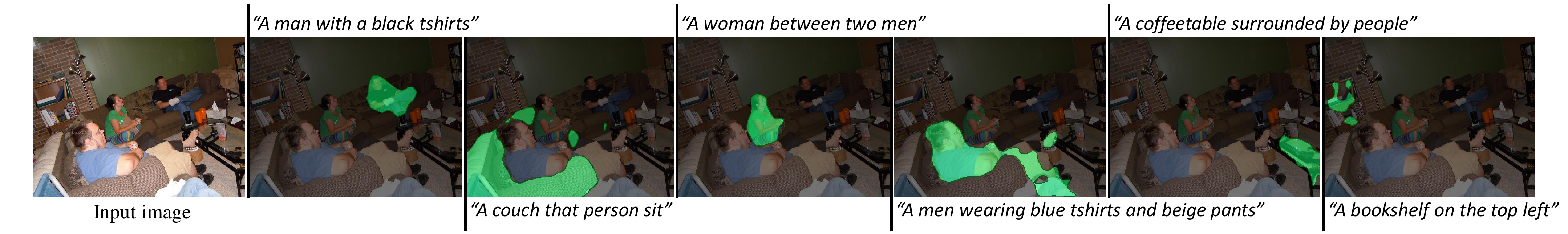}
\vspace{-5mm}
\caption{
Visualization examples of ReSTR according to different language expression queries for an image on Gref \textit{val} set.}
\label{fig:varisent}
\vspace{-2mm}
\end{figure*}

%% file: tables/computation.tex
\begin{table}[t!]
\centering
\scalebox{0.96}{
\small{
\begin{tabular}{@{}l|c|ccc@{}}
\toprule
Methods & DCRF &\# params & {MACs} & {IoU} \\ \midrule
BRINet~\cite{hu2020bi} & \cmark & 241.18M & 367.63G & 48.04 \\
LSCM~\cite{hui2020linguistic} & \cmark & 127.91M & 130.45G & 48.05 \\
CMPC~\cite{huang2020referring} & \cmark & 118.66M & 126.66G & 49.05 \\
ACM~\cite{feng2021encoder} & \xmark & 232.78M & 124.68G & 51.93 \\ \midrule
ReSTR (CME) & \xmark & 122.87M & 52.29G & \textbf{54.48} \\
ReSTR (CME$^{\dagger}$) & \xmark & 108.70M & 52.29G & 54.07 \\ \bottomrule
\end{tabular}
}
}
\vspace{-1mm}
\caption{
Comparison of computations and performance with recent methods.
Both are evaluated on Gref \textit{val} set in IoU (\%).
$\dagger$ denotes the multimodal fusion encoder with weight sharing and MACs is computed with an input image of $320\times 320$.
}
\vspace{-5mm}
\label{tab:computation}
\end{table}

%% file: _5_conclusion.tex
\section{Conclusion} \label{sec:conclusion}

We have proposed ReSTR, the first convolution-free model for referring image segmentation.
ReSTR adopts transformers for both visual and linguistic modalities to capture global context from feature extraction.
It also includes the multimodal fusion encoder composed of transformers to encode fine and flexible interactions between these features of the two modalities.
Also, the multimodal fusion encoder computes an adaptive classifier for patch-level classification.
Furthermore, we have proposed a segmentation decoder to refine the patch-level predictions to the pixel-level prediction in a coarse-to-fine manner.
ReSTR outperformed the existing referring image segmentation techniques on all public benchmarks.
The fact that computational cost quadratically increases as patch size decreases is the potential limitation of our work.
Since the performance of the dense prediction tasks heavily depends on the patch size when using the visual transformer~\cite{strudel2021segmenter}, it introduces an undesirable trade-off between performance and computational cost.
To alleviate this, integrating linear-complexity transformer architectures would be a promising research direction, which we leave for future work.

\vspace{4mm}
{\small
\noindent \textbf{Acknowledgement.} 
We thank Manjin Kim and Sehyun Hwang for fruitful discussions.
This work was supported by 
MSRA Collaborative Research Program, and    %
the NRF~grant and                           %
the IITP grant                              %
funded by Ministry of Science and ICT, Korea
(NRF-2021R1A2C3012728,                      %
 IITP-2020-0-00842,                         %
 No.2019-0-01906 Artificial Intelligence Graduate School Program--POSTECH).                         %
}

%% file: supp/_intro.tex
This supplementary material presents experimental results omitted from the main paper due to the space limit.
\Sec{supp_analy_length} analyzes performance according to language expression length compared with previous methods.
In \Sec{supp_hyperparams}, we investigate the sensitivity of our model to hyperparameters in terms of performance.
Finally, \Sec{supp_more_qual} describes more qualitative results of our method on the Gref dataset.

%% file: supp/_analy_length.tex
\section{Impact of the length of language expression}
\label{sec:supp_analy_length}
\vspace{-1.5mm}
We present the detail analysis of performance according to language expression length in~\Fig{supp_length}.
Following~\cite{liu2017recurrent}, each test set on the dataset is split into four groups in terms of language expression length (\ie~sentence length), and each group is roughly equal size.
Our method outperforms most previous methods except on the 1-5 length group of the Gref dataset, where the gap is marginally 1.2\%p.
Furthermore, our method has less performance degradation from the shortest to the longest sentence length group on four datasets than ACM~\cite{feng2021encoder}, which is the most recent work.
Although ACM is proposed to capture long-range dependencies, it still seems to struggle to understand the complex interaction between words of long language expressions.
Therefore, the performance improvement of ACM mostly comes from its performance on the short sentence length groups.
However, ReSTR shows the improvement of performance on most groups, which suggests that our model captures better long-range interactions of the language expression than the previous work.

%% file: supp/_effect_hyperparam.tex
\section{Sensitivity to hyperparameters}
\label{sec:supp_hyperparams}
\vspace{-1.5mm}
We investigate the effect of the two hyperparameters, the loss balancing weights $\lambda$ and the thresholding value $\tau$, to generate patch-level labels.
The results of our analysis are summarized in~\Fig{supp_hyper}, in which we examine IoU of ReSTR by varying the values of the hyperparameters $\lambda\in \{0.01, 0.05, 0.1, 0.5, 1\}$ and $\tau\in \{0.5, 0.6, 0.7, 0.8, 0.9\}$.
The results suggest that when $\lambda$ is between 0.05 and 0.5, the performance of ReSTR is high and stable, thus insensitive to the hyperparameter setting.
Note that the hyperparameter setting of ReSTR reported in the main paper (the underlined performance in~\Fig{supp_hyper}) is not the best, although it outperforms all existing methods, as we do not tune the hyperparameters to optimize the test performance.

\input{figures/supfig1_leng_analysis}
\input{figures/supfig2_params_search}

%% file: figures/supfig1_leng_analysis.tex
\begin{figure}[t!]
    \centering
    \includegraphics[width=0.99\linewidth]{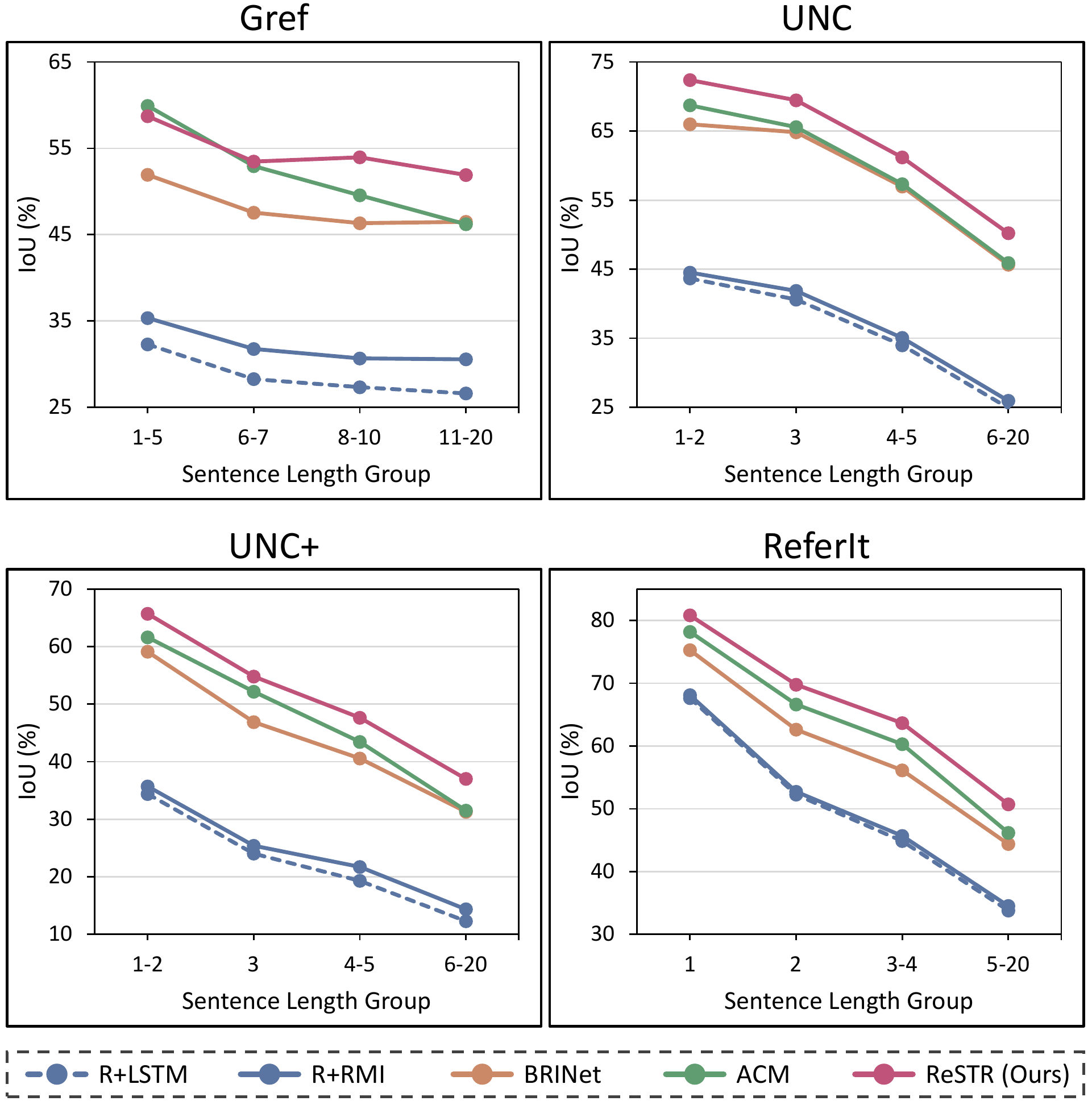}
\vspace{-1mm}
\caption{
Performance in IoU(\%) versus sentence length group on four datasets.
} \label{fig:supp_length}
\vspace{-4mm}
\end{figure}

%% file: figures/supfig2_params_search.tex
\begin{figure}[!t]
    \centering
    \includegraphics[width=0.84\linewidth]{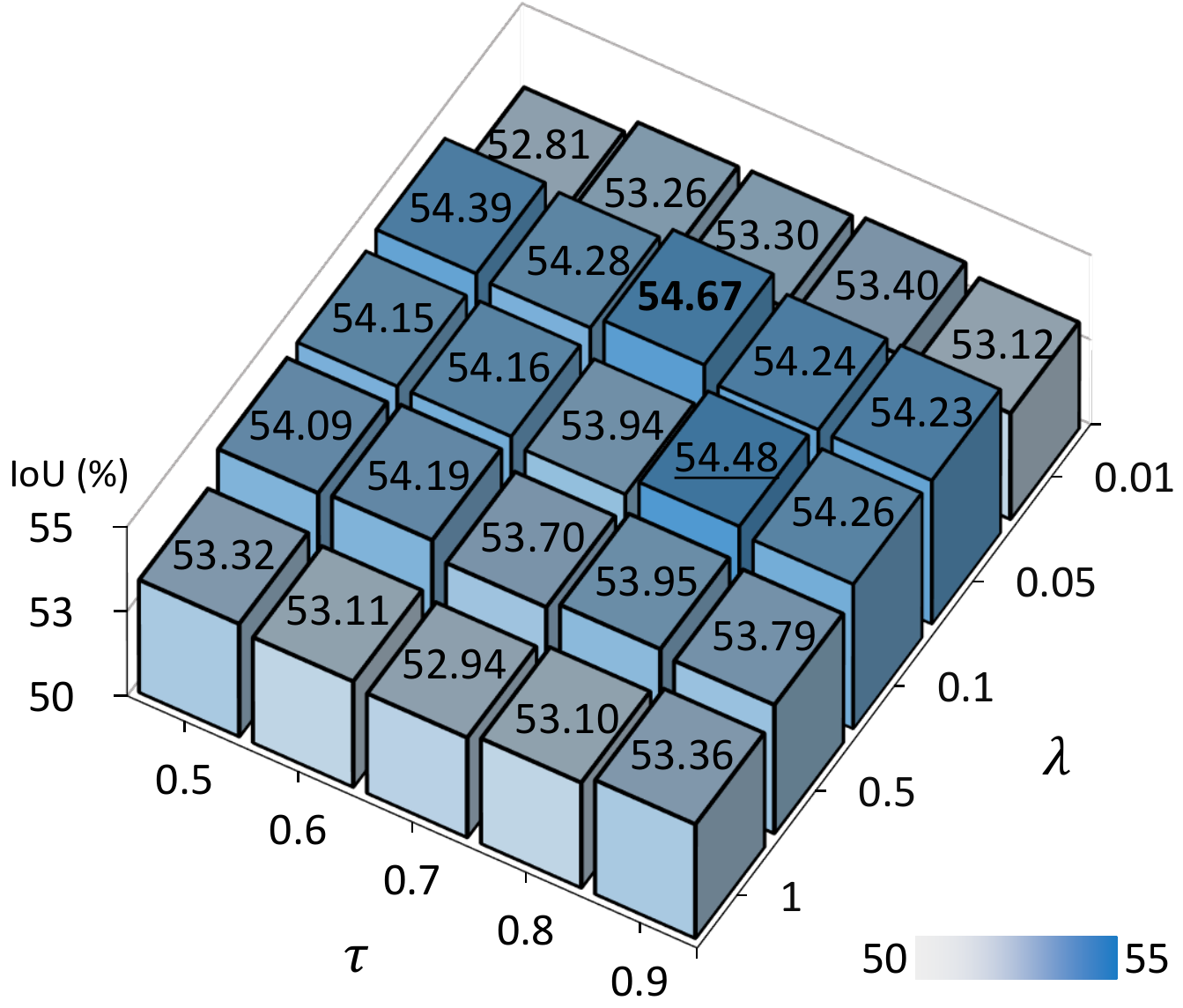}
\vspace{-2mm}
\caption{
Performance in IoU versus $\lambda$ and $\tau$ on the Gref \textit{val} set.
} \label{fig:supp_hyper}
\vspace{-4mm}
\end{figure}

%% file: supp/_more_qual.tex
\section{More qualitative results}
\label{sec:supp_more_qual}
\vspace{-1.5mm}
In~\Fig{supp_qual1} and ~\Fig{supp_qual2}, qualitative results of ReSTR on the Gref dataset are presented.
Pixel-level predictions and the results post-processed with DenseCRF~\cite{Fullycrf} are provided together.
The results show that ReSTR successfully segments masks of the target entities described in various language expressions.
For example, ReSTR predicts accurate masks for language expressions about non-human objects (\Fig{supp_qual1}), partially appeared objects (rows 1-3 in~\Fig{supp_qual2}), and occluded objects (rows 5-7 in~\Fig{supp_qual2}).
Moreover, the qualitative results of the pixel-level prediction show that the segmentation decoder of ReSTR produces the fine-grained prediction as well as removes false positives in the patch-level prediction.
As shown in~\Fig{supp_qual3}, we also present more qualitative results of ReSTR according to varying language expressions for each image.
The results show that ReSTR can comprehend various types of objects (rows 1-2 in~\Fig{supp_qual3}), a sense of locality (rows 3-4 in~\Fig{supp_qual3}), and fine-grained details (row 5 in~\Fig{supp_qual3}).

%% file: figures/supfig3_qual.tex
\begin{figure*}[t!]
    \centering
    \vspace{-1mm}
    \includegraphics[width=0.98\linewidth]{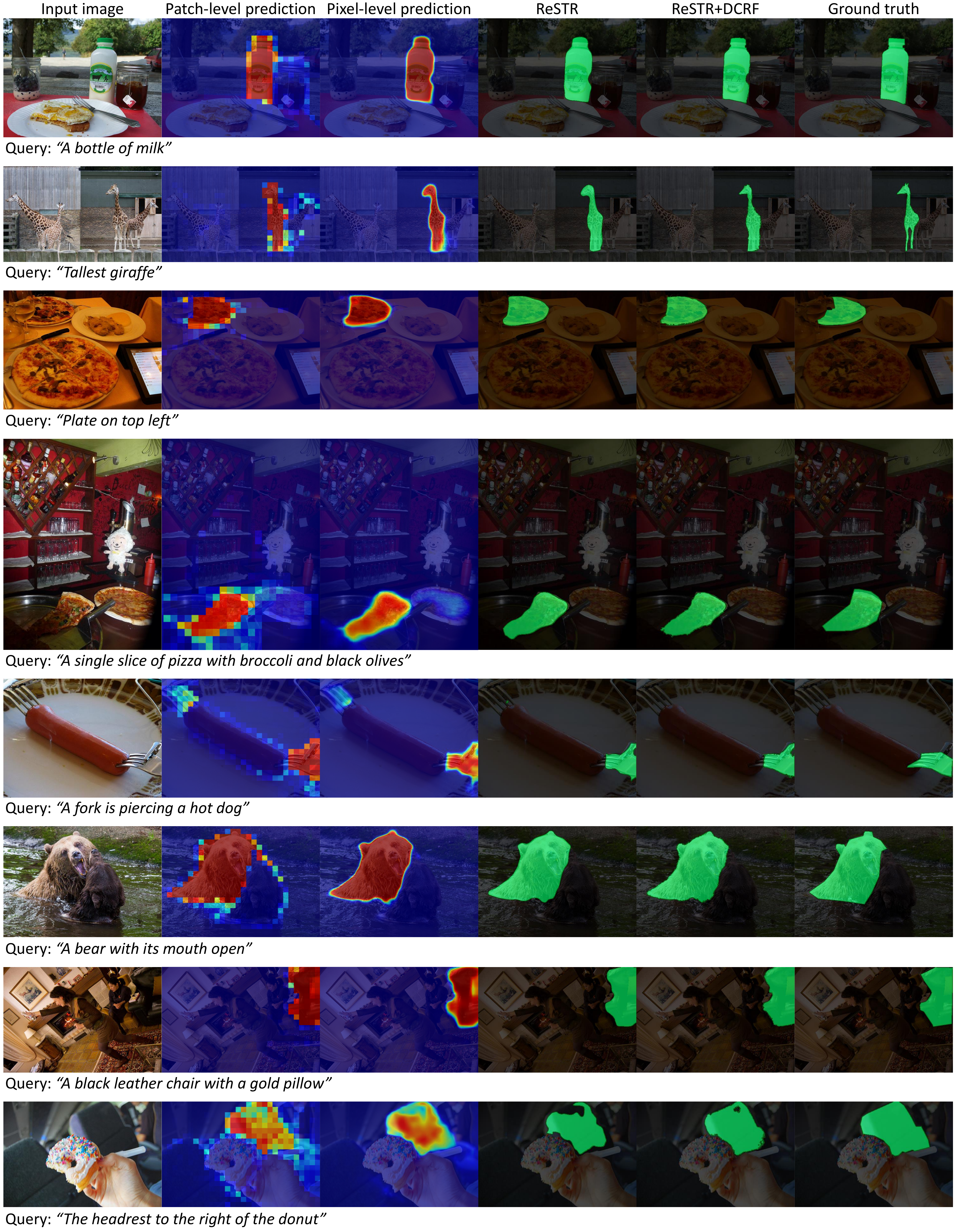}
\caption{
Qualitative results of ReSTR on the Gref \textit{val} set.
} \label{fig:supp_qual1}
\vspace{-2mm}
\end{figure*}

\begin{figure*}[t!]
    \centering
    \vspace{-1mm}
    \includegraphics[width=\linewidth]{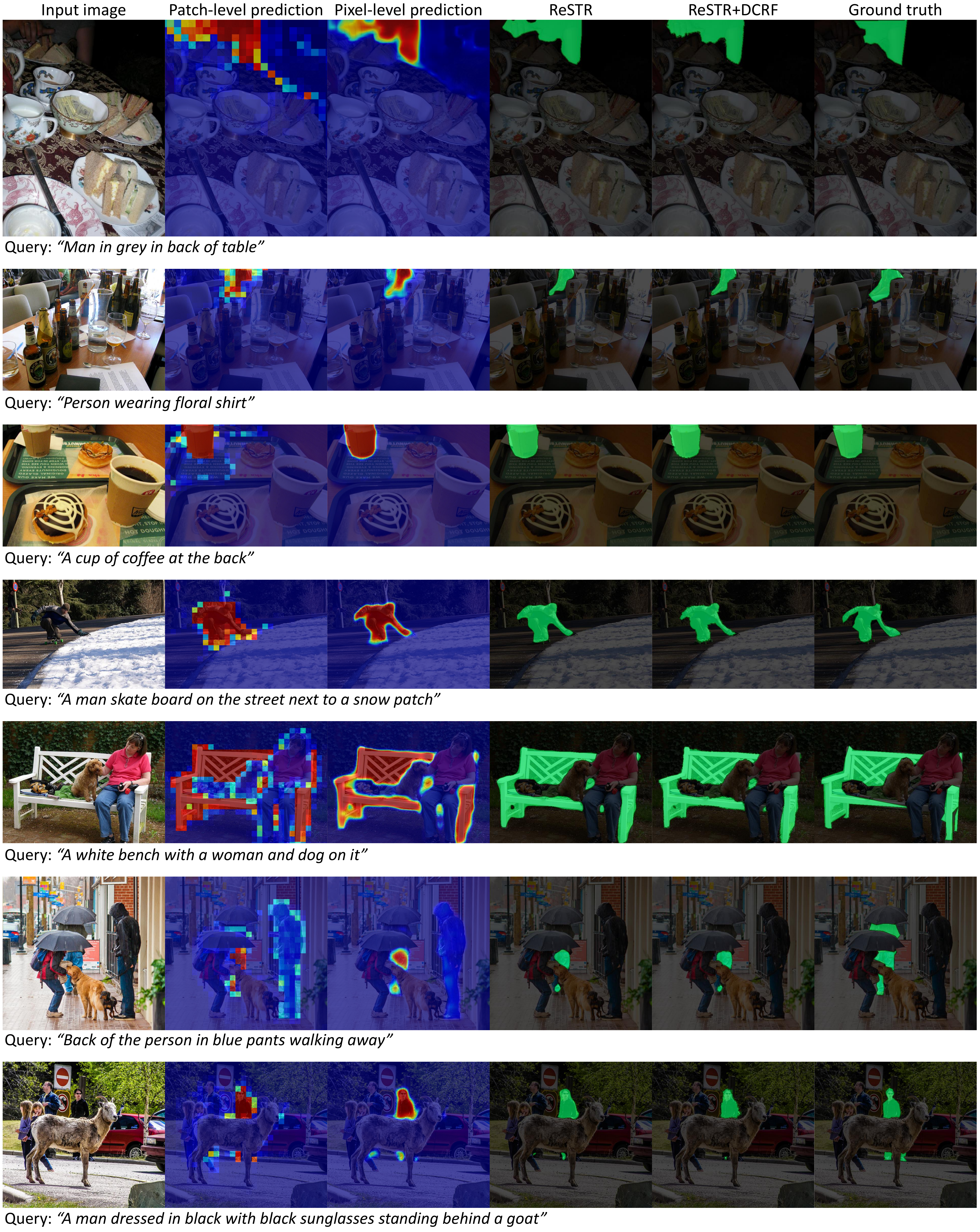}
\caption{
Qualitative results of ReSTR on the Gref \textit{val} set.
} \label{fig:supp_qual2}
\vspace{-2mm}
\end{figure*}

\begin{figure*}[t!]
    \centering
    \vspace{-1mm}
    \includegraphics[width=\linewidth]{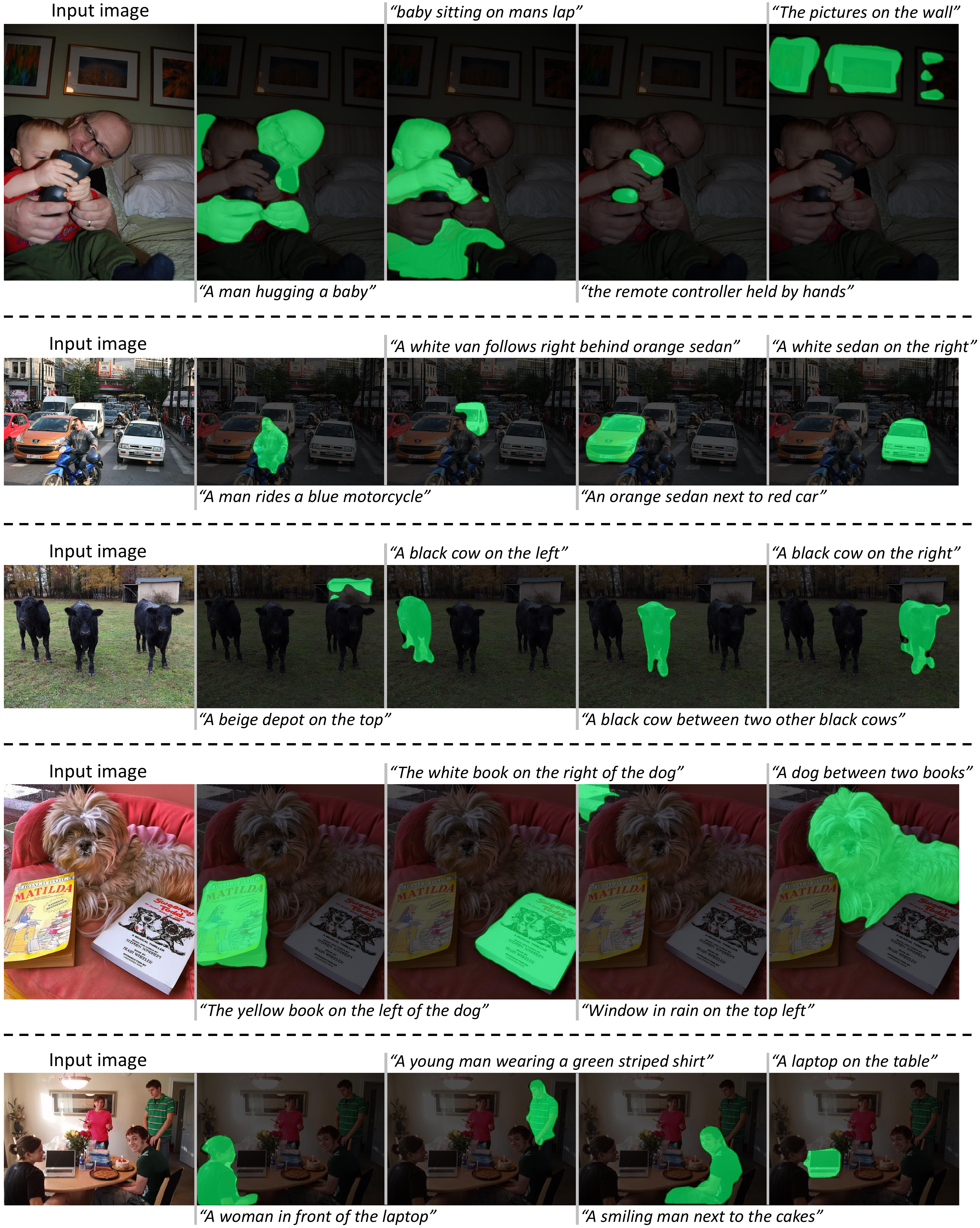}
\caption{
Qualitative results of ReSTR according to different language expression queries for each image on the Gref \textit{val} set.
} \label{fig:supp_qual3}
\vspace{-2mm}
\end{figure*}